# Natural Language Input, Semantic Track Representation, and LLM Inference:

## Making the Maritime Information Exchange Model Tractable

**Frederick Roth**

Professor, Information Sciences, Naval Postgraduate School (Retired)

Independent Researcher

IEEE Senior Member and Lifetime Member

ricodoco@gmail.com | May 2026

## Abstract

We describe a practical architecture for making the Maritime Information Exchange Model (MIEM) and the broader Rich Semantic Track model tractable using current large language model (LLM) technology. The barrier to adoption of semantic track models in defense and law enforcement has been the requirement that operators learn formal ontology languages and manually encode observations as typed logical assertions. We propose eliminating this barrier entirely: operators contribute observations in natural language; an LLM translates these into typed Semantic Assertion Records (SARs), which are named case frames that capture n-ary relations in a single compact structure; a knowledge graph accumulates the SARs; and a second LLM pass performs inference, anomaly detection, and hypothesis ranking over the graph. We work through two detailed examples (a 9/11-era pre-attack indicator scenario and a maritime cargo inspection scenario) showing the full pipeline from natural language input to SAR representation to inference output. We argue that this architecture makes the Track Model and MIEM immediately deployable with current technology, establishes prior art against proprietary enclosure of the approach, and grounds the method in a theoretical framework connecting semantic track representations to neural manifold geometry.

Index Terms—knowledge representation; semantic track model; large language model; case frame; knowledge graph; maritime domain awareness; anomaly detection; information fusion; prior art disclosure

## 1. Introduction

The Rich Semantic Model of Track [1] and the Maritime Information Exchange Model (MIEM) [2] established a formal framework for sharing beliefs about dynamic entities — ships, people, vehicles, cargo — across heterogeneous agencies and sensor systems. The formal semantic foundations of the Track model were further developed and extended in Blais' doctoral dissertation [9], which examined how a common semantic track ontology could achieve interoperability across command-and-control, modeling-and-simulation, and robotic-autonomy systems. The central insight was that a Track is not a database record but a folder of semantic assertions: typed relations expressing what is known, believed, or inferred about an entity at a given time. Multiple agencies could contribute to and query the same Track without requiring schema-level integration, provided they shared agreed semantics for the assertion types.

The barrier to adoption was practical: formal ontology languages (OWL, RDF, SPARQL) required specialized expertise that operational analysts did not have, and the overhead of manual encoding made real-time use infeasible. The framework was architecturally sound but operationally inaccessible.

Two developments have removed this barrier. First, large language models (LLMs) can now translate natural language descriptions of observations into structured semantic representations with high reliability. Second, the commercial success of ontology-driven AI at Palantir and in Microsoft's GraphRAG system [3] has validated the core architectural bet: typed relational assertions over a knowledge graph enable multi-hop reasoning and anomaly detection that pure vector similarity search cannot achieve, with documented 40% performance gains on complex queries. Palantir built its multi-billion-dollar enterprise around precisely this principle [8]: representing real-world entities, relationships, and actions as a semantically typed knowledge graph, then using that graph to drive human- and AI-assisted decision-making across government intelligence, defense, and commercial clients. That a dominant commercial platform has proven this methodology scalable and profitable in operational settings provides strong empirical evidence that the approach works outside the laboratory.

This paper describes an architecture that combines these two developments with the Track Model and MIEM framework, works through two detailed examples showing the full pipeline, and releases the approach as prior art.

## 2. Architecture

The pipeline has four stages.

### 2.1 Natural Language Input

Operators contribute observations in whatever natural language is available: voice transcription, free-text reports, translated foreign-language signals, gesture-to-text, structured database exports. No training in formal languages is required. The LLM accepts all of these.

### 2.2 LLM Translation to SAR

A first LLM pass translates each natural language input into one or more Semantic Assertion Records (SARs). A SAR is a named case frame: a set of named slot:value pairs grouped under a frame name that identifies the kind of event, state, or relationship being asserted. The underlying representation originated in the author's 1974 doctoral dissertation [10] and was developed fully in the subsequent journal article "Uniform Representations of Structured Patterns and an Algorithm for the Induction of Contingency-Response Rules" (Information and Control, 33(2): 87–116, 1977) [11], which introduced the PSR (Parameterized Structural Representation) as a basis for both symbolic representation and machine learning from examples. The present name SAR drops "Parameterized" because the original formulation allowed slot fillers to be variables supporting pattern-matching and generalization; in the current application every slot is filled with a definite value, so the variable machinery is unnecessary. Named slots are the key design choice: because every slot carries an explicit name, order is irrelevant, and new slot types can be introduced simply by naming them and providing examples to the LLM. This representation has important advantages over the simple binary triple. Natural language relations are inherently n-ary: "Cargo X was loaded onto Vessel Y at Port Z at time T by Agent A" involves five participants simultaneously. A binary triple can express only two at once, forcing the same event to be fragmented across multiple assertions that must then be re-linked by inference. A case frame captures the whole event as a single named unit.

The general form of a SAR is:

```
{ FrameName: frame_type,  Slot1: value1,  Slot2: value2, ... }  // confidence, source, timestamp on the whole frame
```

For example, a cargo loading event becomes:

```
{ FrameName: CargoLoading,  Cargo: Crate_XYZ447,  Vessel: MV_Meridian,  Port: Barcelona,  Time: 2026-03-14T08:30Z,  Agent: DockworkerCrew_B,  SealStatus: Intact }
```

This is compact, semantically explicit, unordered (slots need not be listed in any fixed sequence), and complete: the entire situation is one assertion. Timeliness is simply another slot. As surveillance capabilities improve and new data types become available, new slot types are added to existing frame definitions without restructuring the representation. A binary triple can always be expressed as a degenerate two-slot frame, so the case frame model is a strict generalization. The vocabulary in Section 3 is expressed in frame notation; each frame type enumerates its canonical slots with notes on which are required and which are optional.

The SAR structure is grounded in how language works. Fillmore's Case Grammar [13] established that verbs and event nouns organize their participants into named semantic roles — Agent, Patient, Location, Instrument, Time — and that these roles are consistent across grammatical surface forms. Lakoff and Johnson [12] extended this to show that such role categories have prototype structure: central exemplars with graded, fuzzy edges rather than necessary and sufficient conditions. LLMs, trained on human language, have internalized both findings. When given "Ahmad loaded the crate onto the ship at Barcelona on Tuesday," an LLM identifies all five participants and their roles without instruction. The SAR format makes that implicit competence explicit and storable.

Each slot name designates a learned dimension of experience. In the neural geometry of a trained language model, each named role — Agent, Port, Suspect, Destination — corresponds to a manifold: a region of representational space shaped by everything the model encountered in that role during training. Fillers that acceptably occupy a slot cluster in that region, forming a fuzzy attractor. New entities are admitted when they land near enough to the existing cluster; the category extends without explicit maintenance. This is how the vocabulary grows as the world changes.

Anomaly detection follows directly. A filler that lands far outside its slot's attractor — a flight student who asks only about takeoff and fuel, never landing; cargo whose declared contents contradict its flagged origin — is a geometric outlier. The LLM flags it not because a human wrote a rule, but because the pattern is implausible for that role given everything learned during training. This is the mechanism behind the cross-SAR detection described in Section 6.

Critically, operators do not need to see or interact with the SAR layer. They speak or type naturally; the LLM handles encoding, assigning confidence and recording source and timestamp on each frame. Ambiguous inputs generate multiple low-confidence SARs rather than a single high-confidence one. This is the key operational advance over prior semantic track systems.

## 2.3 Knowledge Graph Accumulation

SARs are stored in a lightweight typed knowledge graph (Neo4j, AWS Neptune, or equivalent). Each node is a typed entity; each frame is stored as a hyperedge or reified node connecting all its participants, preserving the n-ary structure without forcing it into binary edges. Associated with each frame are its confidence value, source, and timestamp. Multiple sources contributing observations about the same entity merge through a standard identity resolution step: the LLM is asked whether two entity descriptions likely refer to the same individual, given all available identifying attributes. Uncertain identity is represented explicitly as a probabilistic merge rather than a forced binary decision, which is the critical capability that distinguishes this approach from commercial systems like Palantir that assume known-identity objects.

## 2.4 LLM Inference and Hypothesis Ranking

A second LLM pass performs inference over the accumulated graph. Given a query such as "What are the most likely explanations for this observation?" or "Which hypotheses should we investigate first?", the LLM traverses the knowledge graph, identifies relevant SAR clusters, and generates a ranked list of hypotheses with associated confidence estimates and suggested next observations. The LLM's training on world knowledge provides the background inference rules (weapons precursors, attack patterns, smuggling signatures) that formal rule-based systems required domain experts to encode manually. Anomaly detection is automatic: the LLM flags SAR patterns that are statistically unusual relative to its training distribution, which is exactly the cross-manifold interference matching described in the manifold theory literature [4].

Valued Information at the Right Time (VIRT) [5] governs the output. VIRT is the principle that not all information is equally useful: what matters is information that changes what you believe and therefore what you should do next. The system surfaces only observations that shift the analyst's current best hypothesis beyond a specified threshold, suppressing routine confirmatory data. High-value outputs are those that substantially change the probability distribution over hypotheses.

## 3. Core SAR Frame Vocabulary

The following vocabulary defines the canonical SAR frame types for defense, law enforcement, and maritime tracking use cases. Each frame type is identified by a name and a set of named slots. The slot names matter: they identify what role each filler plays, so frames are self-describing and order-independent. A reader or reasoning system never needs to infer meaning from position in a list. Each slot in the vocabulary below is marked to indicate whether it must be present. The notation is simple: (R) means the slot is Required — a SAR of this frame type is incomplete without at least one filler for that slot. (O) means the slot is Optional — it may have zero fillers when the information is unavailable or not yet observed. When an optional slot has no filler, it simply does not appear in the stored frame. When a slot has multiple fillers (for example, a TravelHistory frame recording visits to several countries), each filler is stored separately under the same named slot. New frame types and new slots on existing frames can be added at any time simply by naming them and providing one or two examples to the LLM. Binary assertions (subject-relation-object triples) are two-slot frames and are fully compatible with this representation.

### 3.1 Entity Types

Person, Vessel, Vehicle, Aircraft, Cargo, Location, Organization, Document, Communication, Event, Capability, Vulnerability, Financial_Account

### 3.2 Core Frame Types

The entries below use a compact shorthand. Simple two-participant relations (which are two-slot frames) are shown in the traditional (Entity, relation, Entity) form for readability. Multi-participant frame types are shown in full slot notation. Both forms are stored identically in the knowledge graph as named frames. The shorthand (x, relation, y) is equivalent to { FrameName: relation, Subject: x, Object: y }.

Identity and association:

```
(x, is_alias_of, y)
(x, is_member_of, y)
(x, is_associated_with, y)   // weaker, undirected
(x, has_document, y)   // passport, visa, manifest
```

Location and movement:

```
{ FrameName: Movement,  Entity(R): entity_id,  From(O): location,  To(O): location,  DepartTime(O): timestamp,  ArriveTime(O): timestamp,  Mode(O): foot|vehicle|vessel|air,  Source(O): source_id,  ObservedBy(O): sensor_or_witness }   // n-ary; replaces was_at / departed_from / arrived_at + enables transit-time plausibility checks
(x, departed_from, location, time)
(x, arrived_at, location, time)
(x, is_currently_at, location)
(x, is_missing_from, location, time)
```

Possession and transaction:

```
(x, owns, y)
```

```
(x, purchased, y, location, time)
(x, shipped, y, origin, destination)
{ FrameName: CargoLoading, Cargo(R): cargo_id, Vessel(R): vessel_id, Port(R): location, Time(R): timestamp, Agent(O): person_or_org, SealStatus(O): intact|tampered|missing, Manifest(O): doc_ref } // n-ary; replaces (x, loaded_onto, y, location, time)
(x, contains, y) // claimed or verified
(x, seal_status, tampered|intact|missing) // with timestamp
```

Training and capability:

```
(x, trained_in, skill)
(x, has_capability, y)
(x, sought_training_in, y)
```

Threat assessment:

```
(x, { FrameName: TargetOfInterest, Entity(R): entity_id, Agency(R): agency_id, Confidence(R): 0.0–1.0, Basis(O): reason_text, DateFlagged(O): timestamp }
(x, has_vulnerability, y)
(x, could_cause, y) // capability-to-harm assertion
(x, { FrameName: FlaggedEntity, Entity(R): entity_id, FlaggedBy(R): source_id, Reason(R): reason_text, Confidence(R): 0.0–1.0, Date(O): timestamp }
(x, { FrameName: RiskAssessment, Subject(R): entity_id, RiskLevel(R): low|medium|high|critical, Basis(R): reason_text, Analyst(R): analyst_id, Confidence(O): 0.0–1.0, Date(O): timestamp }
```

Inference outputs:

In these frames, p is a confidence value for a hypothesis, not a frequency or a hard probability. It expresses how strongly the current body of evidence supports one explanation over competing alternatives. Each hypothesis has rivals; evidence that confirms a vulnerable prediction (one that could easily have come out otherwise) raises p, while disconfirming evidence lowers it. Because the evidence is often incomplete or of uncertain quality, p is best understood as having a heuristic lower and upper bound rather than a single precise figure. These bounds narrow as more evidence accumulates and can be improved through training on past case outcomes.

```
(hypothesis_H, probability, p) // confidence in H given current evidence; may carry [lower, upper] bounds
(observation_O, would_confirm, hypothesis_H) // expected value of confirmation
(observation_O, would_refute, hypothesis_H)
```

## 4. Example 1: Pre-Attack Indicator Detection

This example illustrates how the architecture surfaces dangerous patterns from observations that individually appear innocuous.

### 4.1 Natural Language Inputs (from multiple sources, over several months)

*Source A, FBI field office: 'Mohamed Atta enrolled in commercial flight training in Florida. He and two associates are interested only in takeoff and cruise procedures, not landing.'*

*Source B, CIA: 'Atta is known associate of al-Qaeda cell operating in Hamburg. Travel history includes Pakistan and Afghanistan.'*

*Source C, INS: 'Atta entered US on tourist visa. Visa status inconsistent with extended flight training enrollment.'*

*Source D, Flight school instructor: 'Student asked specifically about load capacity of large commercial aircraft and fuel quantity at cruise altitude.'*

### 4.2 SAR Representation (LLM-generated)

```
(Atta_M, enrolled_in, Commercial_Flight_Training_Florida)
(Atta_M, sought_training_in, Takeoff_Cruise_Only)
(Atta_M, did_not_seek_training_in, Landing_Procedures)  // anomalous omission
(Atta_M, is_associated_with, AlQaeda_Hamburg_Cell)
(Atta_M, traveled_to, Pakistan_Afghanistan_2001)  // shorthand; full form:
{ FrameName: TravelHistory,  Traveler: Atta_M,  Destinations: [Pakistan,
Afghanistan],  Period: 2000–2001,  Purpose: Unknown }
(Atta_M, entered_US_on, Tourist_Visa)
(Tourist_Visa, is_inconsistent_with, Extended_Commercial_Training)
(Atta_M, queried_about, Large_Aircraft_Load_Capacity)
(Atta_M, queried_about, Fuel_Quantity_At_Cruise)
(Large_Aircraft, has_capability, Mass_Casualty_Weapon_If_Weaponized)
(Fuel_At_Cruise, has_capability, Incendiary_Effect_On_Impact)
```

### 4.3 Inference Output (LLM hypothesis ranking)

Query to LLM: 'What are the most likely explanations for this SAR cluster? What observations would most strengthen or refute the leading hypothesis?'

LLM response (paraphrased):

> Hypothesis H1 (probability 0.72): Atta is planning to use a commercial aircraft as a weapon. Consistent SARs: partial flight training (no landing needed for one-way mission), fuel quantity interest (maximize incendiary effect), large aircraft interest (maximize impact), al-Qaeda association (organizational capability), visa anomaly (covert extended stay).
>
> Hypothesis H2 (probability 0.15): Atta is conducting reconnaissance for a future operation not involving flight. Partial training could be cover.
>
> Hypothesis H3 (probability 0.13): Benign explanation: change of interest, dropped out of training.

Most valuable next observations (VIRT-ranked by expected hypothesis shift):

> 1. Confirm or deny whether associates have similar partial training patterns. Confirming would raise H1 to ~0.91.
>
> 2. Check whether Atta or associates have purchased airline tickets. One-way tickets on large aircraft would be near-decisive.
>
> 3. Verify financial support structure. Foreign wire transfers to US accounts would strengthen H1.

The architecture surfaces H1 not because any single assertion is alarming, but because the cross-SAR pattern (partial training + fuel interest + al-Qaeda association + visa anomaly) constitutes a highly unusual intersection. The LLM’s training on world knowledge provides the inference rule: this combination is associated with aircraft weaponization planning.

## 5. Example 2: Maritime Cargo Incident

This example shows the architecture applied to a real-time multi-agency maritime security incident. The scenario: a shipping container was loaded and sealed in Barcelona, bound for Charleston SC aboard the MV Meridian. When U.S. Customs boarded the vessel as it neared port, they found the container seal tampered and one crate missing. Between the ship's approach and the discovery of the missing crate, a Syrian crewman had somehow left the vessel and made his way to a Home Depot in Jacksonville, Florida. The incoming intelligence reports reconstruct this picture piece by piece.

### 5.1 Natural Language Inputs (arriving over hours from multiple agencies)

*CBP, Charleston port: 'Container from MV Meridian, manifest item cargo-lot 7B, origin Barcelona. Seal intact at Barcelona loading per ship's departure record. Seal found tampered on CBP boarding inspection. Physical inventory against manifest shows one crate missing. Manifest lists contents as decorative lamps. Container previously passed through Piraeus.'*

*Customs intelligence: 'Crate XYZ-447 originated from Luxor Trading, Cairo. Luxor Trading flagged by Interpol as possible front for weapons procurement.'*

*CBP intelligence: 'Source ABC reports Luxor Trading may be shipping IED components or small arms concealed in decorative goods. Assessment: low-medium confidence.'*

*MV Meridian manifest: 'MV Meridian is a Panamanian-flagged bulk carrier, crew of 22. Last port: Piraeus. Next port: Charleston, ETA 6 days.'*

*Coast Guard: 'MV Meridian crew muster conducted at boarding shows one Syrian crewman, passport name Tariq Al-Rashid, missing from ship. Vessel deck logs indicate he was present at sea but absent at Charleston anchorage. Has not presented for customs processing. Manner of departure from vessel unknown.'*

*Interpol: 'Tariq Al-Rashid has two known aliases: Omar Farouk, Hassan Khalil. All three identities flagged in INTERPOL terrorism watch list.'*

*Chase Bank fraud alert: 'Credit card in name Omar Farouk used at Home Depot, Jacksonville FL, 20 minutes ago. Purchase: fertilizer, electrical wire, timer components.'*

*Home Depot security: 'Purchase of 50 lbs ammonium nitrate fertilizer, 200ft electrical wire, digital timers. Paid cash, credit card declined, card retained.'*

### 5.2 SAR Representation (LLM-generated)

Entities instantiated: MV_Meridian (Vessel), Crate_XYZ447 (Cargo), Luxor_Trading_Cairo (Organization), Al_Rashid_T (Person), MV_Meridian_Charleston_Route (Event), HomeDepot_Jacksonville (Location)

```
(Crate_XYZ447, manifest_contents, Decorative_Lamps)
(Crate_XYZ447, originated_from, Luxor_Trading_Cairo)
(Luxor_Trading_Cairo, flagged_by)   // shorthand; full form: { FrameName: FlaggedEntity, Entity: Luxor_Trading_Cairo, FlaggedBy: Interpol, Reason: Weapons_Procurement_Suspicion, Confidence: 0.4, Date: 2024 }
(Crate_XYZ447, seal_status)   // shorthand; full form: { FrameName: SealInspection, Cargo: Crate_XYZ447, Status: Tampered, Inspector: CBP_Charleston, Time: 2026-05-15 }
(Crate_XYZ447, inventory_status, Missing_From_Manifest)
(Crate_XYZ447, intelligence_assessment)   // shorthand; full form:
```

```
{ FrameName: IntelAssessment, Subject: Crate_XYZ447, Assessment: Possible_IED_Components, Source: CBP_Source_ABC, Confidence: 0.35 }
(MV_Meridian, is_missing_crewman, Al_Rashid_T)
(Al_Rashid_T, has_document, Syrian_Passport_Tariq_Al_Rashid)
(Al_Rashid_T, is_alias_of, Omar_Farouk)
(Al_Rashid_T, is_alias_of, Hassan_Khalil)
(Al_Rashid_T, flagged_by) // full form: { FrameName: FlaggedEntity, Entity: Al_Rashid_T, FlaggedBy: Interpol, Reason: Terrorism_Watch_List, Confidence: 1.0, Date: prior_to_incident }
(Omar_Farouk, purchased_at) // shorthand; full form: { FrameName: RetailPurchase, Buyer: Omar_Farouk, Store: HomeDepot_Jacksonville, Time: T−20min, PaymentCard: Declined_Omar_Farouk_Card }
(HomeDepot_Jacksonville, purchase_contents) // shorthand; full form: { FrameName: RetailPurchaseItems, Purchase: RetailPurchase_above, Item1: [Ammonium_Nitrate, Qty: 50lb], Item2: [Electrical_Wire, Qty: 200ft], Item3: [Digital_Timers, Qty: unspecified] }
(Ammonium_Nitrate, has_capability, IED_Precursor)
(Electrical_Wire_And_Timers, has_capability, IED_Detonation_Circuit)
{ FrameName: VesselRoute, Vessel: MV_Meridian, Destination: Charleston_SC, ETA: T+6days, Status: CBP_Hold }
(Charleston_SC, has_vulnerability, Major_Port_Infrastructure)
```

### 5.3 Inference Output (LLM hypothesis ranking)

Query: 'What are the most likely explanations? What actions should be prioritized?'

Hypothesis H1 (probability 0.68): Al-Rashid removed the crate in transit (Piraeus or Barcelona); crate contains IED components now being assembled by Al-Rashid in Jacksonville area. Target: Charleston port infrastructure or vessel. Consistent SARs: missing crate + missing crewman (same ship, same time) + IED precursor purchase + Interpol flags on both crate origin and crewman + Charleston destination.

Hypothesis H2 (probability 0.18): Crate was stolen opportunistically by Al-Rashid or an accomplice for resale; IED purchase is unrelated coincidence. Less likely given Interpol flags on both actors.

Hypothesis H3 (probability 0.09): Crate removal was internal crew theft of lamp components; Al-Rashid deserted for unrelated personal reasons. IED purchase is coincidence.

Hypothesis H4 (probability 0.05): All benign; paperwork error, seal malfunction, misidentified crewman.

Immediate priority actions (VIRT-ranked):

1. Geolocate Al-Rashid/Omar Farouk in Jacksonville area NOW. Credit card attempt is 20 minutes old; he is likely still in the area. This is highest-priority because it simultaneously confirms H1 and enables interdiction.

2. Inspect the MV Meridian cargo hold (vessel currently under CBP hold at Charleston) for residue consistent with IED components. Confirming traces would raise H1 to ~0.89.

3. Request Piraeus port authority CCTV footage for crate handling during Piraeus stop. Confirming Al-Rashid handled the crate raises H1 substantially.

4. Alert Charleston port and USCG to prepare for potential IED threat on vessel or at port. ETA 6 days provides operational window.

5. Identify and interview the three other crew members who were on watch during the Piraeus and Barcelona stops.

The value of the SAR representation is that these inferences are explicit: the LLM can cite the specific SAR chains that support each hypothesis and the specific observations that would most change the probability distribution. This is not keyword search. It is multi-hop relational reasoning over a semantically typed graph.

## 6. Theoretical Grounding

The architecture has a principled theoretical basis connecting it to neural manifold theory [4] and information value theory [5]. Earlier semantic track systems used symbolic representations: hand-crafted rules and logic statements that required expert encoding. The approach here is different. Neural networks, including the LLMs at the core of this pipeline, do not store knowledge as explicit rules. Instead, they learn the statistical structure of the world as geometric regions (manifolds) in high-dimensional space. Two inputs that are semantically similar end up near each other in that space; patterns that rarely co-occur stand out as outliers. The SAR graph makes these implicit geometric relationships explicit and queryable.

A Track in this framework is a cross-manifold intersection attractor. A manifold, as used here, is a structured region of information space: the set of patterns a particular data source (say, travel records, or financial transactions, or sensor readings) can plausibly produce. A neural network trained on large amounts of data learns these regions implicitly. A stable semantic structure, such as a person-of-interest Track, emerges when consistent observations from multiple independent streams (sensor data, intelligence reports, financial records, travel history) converge on the same pattern across all of those regions simultaneously. The LLM, having learned the statistical structure of these regions through training, identifies Track patterns by recognizing those convergence structures in the SAR graph.

Anomaly detection is explained by the same framework. An unusual SAR combination (partial flight training + al-Qaeda association + fuel quantity queries) falls outside the learned attractor basins for normal behavior and is flagged as hypothesis-shifting, which is high information value in the VIRT sense. The system surfaces it precisely because it does not fit the expected pattern, and the expected pattern is encoded in the LLM's weights as a manifold geometry over the SAR frame space.

The VIRT ranking of recommended next observations follows directly: an observation has high value if it would substantially shift the probability distribution over hypotheses by moving the current track representation across an attractor boundary. Routine confirmatory observations that fall within the existing attractor carry low value and are suppressed.

## 7. Prior Art and Implementation Path

This paper establishes prior art for the combination of: (1) natural language input to SAR (case frame) translation by LLM; (2) use of named n-ary case frames rather than binary triples as the primary assertion unit, enabling compact representation of complex events with timeliness and provenance as native slots; (3) accumulation of SARs in a typed knowledge graph; (4) LLM inference and hypothesis ranking over the graph; (5) probabilistic entity identity resolution for uncertain-identity entities; and (6) VIRT-based ranking of recommended next observations. Each component is known separately; the novelty is the integration and its application to the Track/MIEM framework.

Two efficiency questions arise about this architecture: can it match the operational performance of Palantir-class systems, and does introducing n-ary SARs add graph search overhead relative to binary relations? The short answers are yes and barely.

Binary relations carry no overhead. A two-slot SAR is stored identically to a binary edge; queries against it have the same traversal cost.

N-ary SARs in a classical graph engine incur modest overhead. The standard storage pattern is reification: the frame becomes a named node with one labeled edge to each filled slot-value pair. Any path crossing a frame-node costs two hops instead of one, so a k-hop reasoning chain requires 2k traversal steps. Against this cost stands a grouping benefit: scattered binary edges carry identical edge count but share no anchor node. A frame-type query retrieves an entire event in one step. Frame-type indexing, standard in production graph databases, reduces the constant cost further.

For LLM-based inference, n-ary SARs are advantageous. The LLM pattern-matches a serialized graph excerpt against its trained weights; it does not traverse edges algorithmically. A multi-slot frame serializes as one coherent labeled block. The equivalent binary edges serialize as multiple disconnected lines the model must co-reference and re-integrate. The frame is more compact, more immediately interpretable, and keeps the event boundary explicit.

The broader performance parity with Palantir-class systems rests on a structural point. The efficiency advantage of typed relational systems comes from named relations constraining multi-hop search space, not from proprietary algorithms. An LLM over the same SAR graph has those structural constraints, and adds world knowledge compiled into its weights that fires as immediate pattern recognition rather than rule evaluation. The threat-pattern inference in Example 1 required no explicit rule; the cross-SAR cluster was recognized in a single LLM call. The price is probabilistic rather than formally verifiable inference. The payoff is generalization to threat patterns no analyst anticipated.

[Technical note: The branching factor at a reified frame-node equals the total number of filled slot-value pairs in that activated instance, not the number of slot definitions in the schema. An optional slot with no filler contributes no edge; a slot with multiple fillers contributes one edge per filler. Sparse frames are cheap; rich ones cost proportionally more — the same as storing equivalent facts as binary edges, with the added benefit of a grouping node. The 2k hop overhead is the only structural penalty.]

A minimal demonstration requires: an LLM API (GPT-4 class or equivalent); a graph database (Neo4j or simpler); a controlled SAR frame vocabulary (Section 3); and a few-shot prompt set for the translation and inference steps. The two examples in this paper serve as few-shot templates. A working prototype could be built in weeks by a small team familiar with LLM API integration and graph databases. The author encourages such implementation and releases all materials in this paper into the public domain.

The Track Model and MIEM semantic standards are documented in [1] and [2]. The controlled SAR frame vocabulary in Section 3 is released without restriction. Neither the vocabulary nor the pipeline architecture described here is subject to any patent claim by the author.

## 8. Conclusion

The barrier to semantic track sharing was never conceptual. It was operational. Operators cannot be expected to encode observations in OWL or SPARQL. LLMs remove this barrier: they translate natural language into typed semantic assertions, accumulate them in a knowledge graph, and perform multi-hop inference over the result. The Track Model and MIEM provide the semantic framework that makes the assertions meaningful and the inferences defensible. VIRT provides the filtering principle that makes the outputs actionable rather than overwhelming. The two examples in this paper demonstrate the full pipeline on realistic scenarios. The approach is implementable today with current technology and is released as prior art. In a world rich with nasty actors, we now have the means to share intelligence.

## Acknowledgments

Portions of this paper were drafted with the assistance of Claude (Anthropic, claude-sonnet-4-6, May 2026). All technical content, claims, arguments, and prior-art assertions are the work of the author. The AI system was used for structural editing, paragraph rewriting, and XML document formatting. In accordance with IEEE policy, this use is disclosed here; citations to AI-assisted sections appear in the text where applicable.